# SAAN: Similarity-aware attention flow network for change detection with VHR remote sensing images

Haonan Guo, *Student Member, IEEE*, Xin Su, *Member, IEEE*, Chen Wu, *Member, IEEE*, Bo Du, *Senior Member, IEEE*, and Liangpei Zhang, *Fellow, IEEE*.

*Abstract*—Change detection (CD) is a fundamental and important task for monitoring the land surface dynamics in the earth observation field. Existing deep learning-based CD methods typically extract bi-temporal image features using a weight-sharing Siamese encoder network and identify change regions using a decoder network. These CD methods, however, still perform far from satisfactorily as we observe that 1) deep encoder layers focus on irrelevant background regions and 2) the models' confidence in the change regions is inconsistent at different decoder stages. The first problem is because deep encoder layers cannot effectively learn from imbalanced change categories using the sole output supervision, while the second problem is attributed to the lack of explicit semantic consistency preservation. To address these issues, we design a novel similarity-aware attention flow network (SAAN). SAAN incorporates a similarity-guided attention flow module with deeply supervised similarity optimization to achieve effective change detection. Specifically, we counter the first issue by explicitly guiding deep encoder layers to discover semantic relations from bi-temporal input images using deeply supervised similarity optimization. The extracted features are optimized to be semantically similar in the unchanged regions and dissimilar in the changing regions. The second drawback can be alleviated by the proposed similarity-guided attention flow module, which incorporates similarity-guided attention modules and attention flow mechanisms to guide the model to focus on discriminative channels and regions. We evaluated the effectiveness and generalization ability of the proposed method by conducting experiments on a wide range of CD tasks. The experimental results demonstrate that our method achieves excellent performance on several CD tasks, with discriminative features and semantic consistency preserved.

## I. Introduction

Change detection (CD) aims to identify different states of objects and phenomena by analyzing two or more remote sensing images of the same area taken at different times. Due to the increasing need for timely monitoring of the dynamics on the earth's surface, CD has become an indispensable technique for various domains, such as ecosystem monitoring[1], emergency response[2], and agricultural management[3].

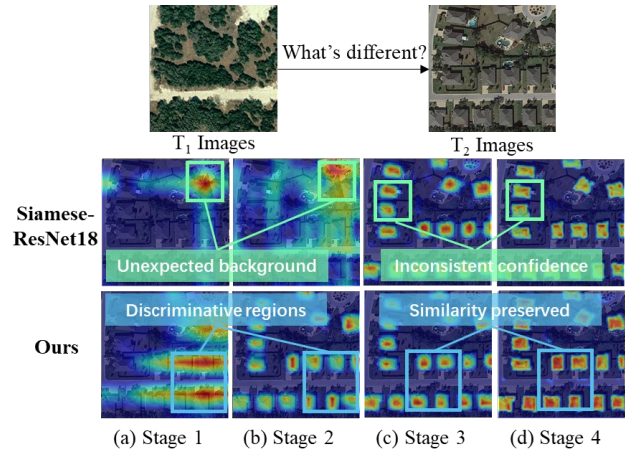

Fig.1. Illustration of the drawbacks of existing Siamese neural network architecture. Typically, this model focuses on unexpected background in the deep features, and the model's confidence in the change regions is inconsistent at different stages. In comparison, our proposed method can focus on discriminative change regions with the similarity relation preserved at all stages.

Early stages of CD studies focused on identifying change information from mid- and low-resolution remote sensing images using manually-designed criteria[4], such as image algebras[5], [6], mathematical transformations[7], [8], and post-classification comparisons[9]. However, these handcrafted criteria were developed for specific areas and may not perform well in large-scale CD applications. Moreover, traditional CD methods developed for mid- and low-resolution remote sensing images can only provide a coarse view of the change region[10] and cannot meet the requirement of refined dynamic monitoring. Although the very high-resolution (VHR) images provide a more ideal data source for refined CD, handcrafted feature-based methods usually cannot generalize well on VHR images that contain more high-frequency information and intra-object variety[11]. As a result, there is a need for more advanced approaches to improve the accuracy of CD on VHR images.

Corresponding authors: Chen Wu and Xin Su.
Haonan Guo, Chen Wu and Liangpei Zhang are with the State Key Laboratory of Information Engineering in Surveying, Mapping and Remote Sensing, Wuhan University, Wuhan, China (e-mail: haonan.guo@whu.edu.cn; chen.wu@whu.edu.cn; zlp62@whu.edu.cn).

Xin Su is with the School of Remote Sensing and Information Engineering, Wuhan University, Wuhan, China (e-mail: xinsu.rs@whu.edu.cn).
Bo Du is with the National Engineering Research Center for Multimedia Software, Institute of Artificial Intelligence, School of Computer Science and Hubei Key Laboratory of Multimedia and Network Communication Engineering, Wuhan University, Wuhan, China (e-mail: gunspace@163.com).

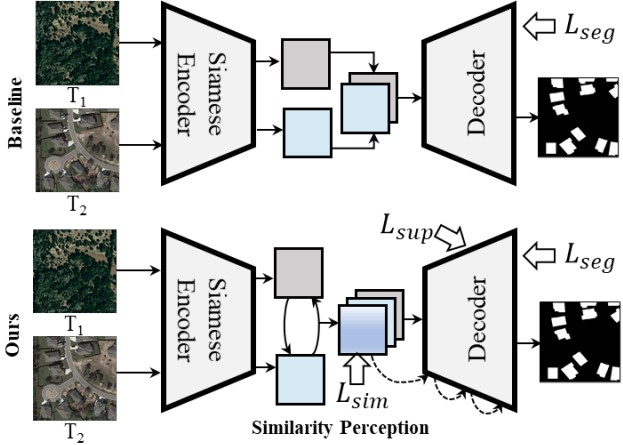

Fig.2. The motivation of our proposed SAAN framework.

Over the past decade, the flourishing deep learning (DL) technique has been applied to VHR image change detection tasks and has successfully shown its superiority over traditional methods[12]. Prevailing DL-based CD methods usually adopt the Siamese neural network architecture[13] to extract temporal image features with a weight-sharing encoder network, followed by a decoder network that refines feature resolution and identifies change regions[12]. The commonly adopted encoder network includes ResNet[14], SE-Net[15], and VGGNet[16] which are designed for natural images. Intuitively, deep features extracted from these encoder networks are expected to contain richer semantic information and be more discriminative than shallow-layer features. However, as depicted in Fig.1, we surprisingly observe that the decoder struggles to distinguish change regions from deep encoder features in the early decoder stage. Instead, the decoder focuses on irrelevant background noises that deteriorate CD performance. Moreover, the model's confidence in the changing areas fluctuates at different decoder stages. These problems are caused by two typical drawbacks of this widely adopted Siamese network architecture, which are described as follows.

1) Previous studies have highlighted the significance of high-quality extracted features in achieving satisfactory performance in CD[17], [18]. However, many existing methods merely optimize their models by the loss calculated at the highest feature resolution in the decoder stage. This approach hampers the ability of deep convolutional layers to learn representative image features due to the vanishing gradient problem caused by unbalanced categories, where merely 0.5% of the land surface is changing annually[19]. 2)The second problem, as illustrated in Fig.1, is the inconsistency in the model's attention towards the change region across different decoder stages, rather than maintaining a higher level of confidence at the higher decoder stages. This inconsistency arises from the presence of redundant non-changing noises. Features from shallow convolutional layers contain general texture and edge information of the whole image. However, the decoder fails to suppress the irrelevant information from the non-changing region since it cannot perceive the correlation between bi-temporal image features.

To tackle the first problem, we argue that features extracted from the Siamese encoder network should exhibit semantic dissimilarity in changing regions and similarity in unchanged regions within the feature space. To achieve this, we introduce a similarity optimization strategy in the encoder stage to guide the model to explicitly learn similarity correlation in the changing and non-changing regions, as illustrated in Fig.2. To mitigate the second drawback, we propose a simple yet effective similarity-guided attention flow module that utilizes feature similarity relation as attention to help the model progressively focus on the changing regions to avoid the interference of non-changing noises. Furthermore, we enhance the model's perception of feature similarity across different decoder stages by designing an attention flow mechanism to interconnect adjacent attention blocks. This mechanism enables the flow of the similarity information and promotes effective communication between decoder stages. By incorporating similarity optimization and the similarity-guided attention flow mechanism, the prevailing Siamese neural network architecture can be improved by extracting semantically informative deep encoder features while maintaining higher confidence in the decoder network with the similarity relation preserved, as illustrated in Fig.1.

To validate the effectiveness and generalization capability of our proposed approach, we conducted extensive experiments on three CD tasks: building change detection, general change detection, and semantic change detection. Our method explicitly optimizes the similarity correlation in the deep feature space and introduces similarity-guided attention flow blocks to preserve similarity semantic consistency. By optimizing the similarity relation, our encoder network can learn discriminative features, which, in turn, guide the decoder to focus specifically on the changing areas. The experimental results show that our method significantly improves CD accuracy across various CD tasks. Overall, the contributions of this paper can be summarized as follows:

1. A similarity-guided attention flow mechanism is proposed to generate spatial-channel attention maps under the guidance of feature similarity relations. By leveraging these attention maps, change feature representation can be enhanced while the irrelevant non-changing information can be suppressed. To address the challenge of feature similarity inconsistency across different stages, we employ attention flow mechanisms to enable the flow of similarity relations at various decoder stages, thereby preserving the model's perception of feature similarity relations. This helps maintain the model's awareness of feature similarity relation in all the decoder stages.
2. We introduce a similarity optimization strategy in the Siamese encoder-decoder architecture to encourage the encoder to discover the semantic correlation between bi-temporal images. By explicitly optimizing feature similarity relation in the feature space, the extracted features are semantically similar in the unchanged regions while exhibiting differences in the changing regions. This optimization approach allows the extraction of high-quality features that contribute to the distinction of discriminative change regions in the decoder stage.

3. By incorporating similarity optimization and the similarity-guided attention flow module, we propose a novel change detection framework named SAAN. This framework aims to accurately identify change regions in bi-temporal images by improving the quality of extracted features and enhancing the representation of change-related information. This leads to more precise and reliable change detection results. Moreover, this architecture can be integrated into the existing CD methods and improve the accuracy of various CD tasks.

The rest of this paper is organized as follows. Section II presents an in-depth overview of related works in the field of change detection (CD). Our proposed method is described in Section III. Implementation details and analysis of the experimental results are provided in Section IV. Conclusions are presented in Section V.

## II. Related Works

Remote sensing image change detection (CD) is a task that involves identifying changes in objects and phenomena between two images of the same scene captured at different times [11], [20]. According to the practical demand for CD, CD can be categorized into general change detection which focuses on all elements of the ground object change, change detection that is interested in specific types of ground objects (e.g., buildings), and semantic change detection that exploits 'from-to' change information from bi-temporal images. Despite different change types of interest, the purpose of CD is to exploit and identify temporal differences from bi-temporal images. In the following subsections, we introduce these CD methods, categorized by their development stages.

### A. Early Change Detection Methods

Limited by imaging technology, early CD methods were developed primarily based on medium- and low-resolution remote sensing images, which merely provide a macro view of the land surface[21]. These methods can be categorized into algebra-based, transformation-based, and classification-based methods. Algebra-based methods aim to establish criteria, such as differing[5] and rationing[22], regression relations [23], and change vectors[6], to exploit change information from multi-temporal remote sensing images. However, since the change information is detected from raw images, algebra-based methods rely heavily on the assumption that multitemporal images are highly aligned in radiation, which does not hold in practical scenarios where imaging conditions, including radiation, atmosphere, and viewing angles, may vary across different acquisition times[11].

To alleviate this limitation, transformation-based methods such as principal component analysis[24] and multivariate alteration detection[7], seek to transform the input multitemporal images into a shared feature space. In this feature space, the areas that remain unchanged exhibit high correlation, enabling the highlighting of changing areas[8]. However, these methods struggle to perform well on VHR images that contain more detailed high-frequency information[25], as they consider only spectral information and neglect the spatial context relation that is crucial for VHR image interpretation.

Apart from detecting whether changes occur within regions of interest, it is also important to ascertain the specific categories of change types [9]. To identify this "from-to" change type, the classification-based methods aim to predict the complete change matrix that represents all types of land-use or land-cover alterations[26]. These methods generate the "from-to" change map by comparing the two classification maps derived from bi-temporal images. However, the risk of accumulated classification error places high demands on accurately classifying both images[27]. Since these traditional methods were originally designed for medium- and low-resolution images, their performance is limited when applied to VHR images that encompass greater high-frequency information and intra-object variability.

### B. Deep learning-based methods

With the success of deep learning in the computer vision field, deep learning-based CD approaches have demonstrated superior efficacy compared to traditional methods since they can fully leverage spatial-spectral information and automatically learn from VHR images[12]. In the following subsections, we will introduce the prominent deep learning-based CD techniques, encompassing segmentation-based methods, metric learning-based methods, and attention mechanisms specifically designed for CD.

*1) Segmentation-based Change Detection*

A prevailing series of deep learning-based CD methods are based on the Siamese neural network. Different from the common semantic segmentation task that takes a single image as input, the CD task takes images captured at different times as input. To adapt the input of the CD task, some early methods attempted to directly concatenate the bi-temporal images before feeding into the network[28], which are prone to omit the small change areas in VHR images[29]. To effectively and efficiently detect changes within VHR images, the Siamese encoder-decoder architecture has been designed[30]. This architecture leverages a weight-sharing encoder network to extract temporal image features and employs a decoder network to generate the change map. Building upon this architecture, Daudt et al. introduced two variants of the Siamese network, namely FC-siam-conc, which concatenates bi-temporal image features in the decoder stage, and FC-siam-diff, which calculates the absolute difference of the bi-temporal features in the decoder stage [31]. Subsequent studies have focused on enhancing CD performance from the perspectives of feature extraction and feature discrimination. To improve the representation of bi-temporal features, many studies have adopted backbones that are pretrained on ImageNet such as VGG-Net[32], Res-Net[20], and SE-Net[33] rather than training the model from scratch. Zhang et al. introduced a deep supervision strategy in the decoder part to enhance bi-temporal feature learning[32]. Liu further enhanced remote sensing image feature representation by employing an encoder network pre-trained on remote sensing tasks[34]. Driven by the outstanding performance of the Transformer network, Zhang et al.[35] introduced the transformer mechanism in the encoder and decoder parts to enhance change feature representation. Furthermore, to better exploit change information from the extracted features, certain studies have improved the decoder part by recurrent neural networks[30], attention mechanisms[36], and boundary refinements[37]. For instance, Zhang et al.[20] integrated a superpixel sampling network[38] into the Siamese U-Net architecture, successfully suppressing latent noise and refining edge integrity. Despite these endeavors to enhance the Siamese encoder-decoder architecture, these segmentation-based methods are inadequate in effectively leveraging change information by implicitly learning the discrepancy between predictions and the corresponding ground truth at the highest feature resolution.

*2) Mining similarity relation*

Different from segmentation-based methods, metric learning-based methods adopt an explicit approach to optimize feature similarity between bi-temporal images at the pixel or region level[39]. Originally developed for dimension reduction purposes, the similarity optimization strategy aims to project high-dimensional data onto a low-dimensional output space, wherein "similar" points are mapped in close proximity on the manifold, typically through the utilization of contrastive loss functions[40]. This loss function minimizes the parameterized distance between features from similar object pairs while maximizing it for dissimilar object pairs. The efficacy of contrastive loss has been demonstrated in the field of face recognition, as features representing the same individual's face exhibit semantic similarity in the metric space[41]. The further improvements mainly lie in the design of advanced parameterized distance[42] or optimization objectives [43] aimed at enhancing feature discrimination[44]. It's worth noting that the effectiveness of the similarity optimization strategy extends beyond face recognition and finds applications in diverse fields, including self-supervised learning[45], image retrieval[46], and multimodal alignment[47].

The motivation for applying similarity optimization in CD is to learn a feature space where the pairs of feature vectors representing changed regions exhibit substantial dissimilarity, while pairs representing unchanged regions demonstrate proximity to each other[48]. By explicitly optimizing the distance between bi-temporal features extracted from the Siamese CNN, the efficacy of feature representation is enhanced, thus facilitating the extraction of time-series correlations. For example, Zhang et al.[49] introduced a weighted contrastive loss[40] in a Siamese convolutional neural network to explicitly optimize the distance between bi-temporal image features; a threshold is then used to determine the change areas by binarizing the Euclidean distance map in the test phase. Building upon the contrastive loss, Zhang et al. proposed a triplet loss to capture the semantic correlation among features extracted from Siamese CNN[50], consequently improving the interclass separability and intraclass consistency of features. As continuous down-sampling operations result in the loss of spatial details in input images, Chen and Shi [48] further enhanced feature representation by integrating features of different scales and introducing self-attention mechanisms to exploit spatial and temporal relations. Furthermore, Shi et al. applied deep supervision and soft attention mechanisms to further improve the feature representation in CD[51]. Despite the considerable feature representation capability and interpretability of metric learning-based CD methods, the obtained CD results are likely to perform badly in the misregistered regions where the features are misaligned. Incorporating metric learning-based methods into segmentation-based methods may complement the advantages of both methods.

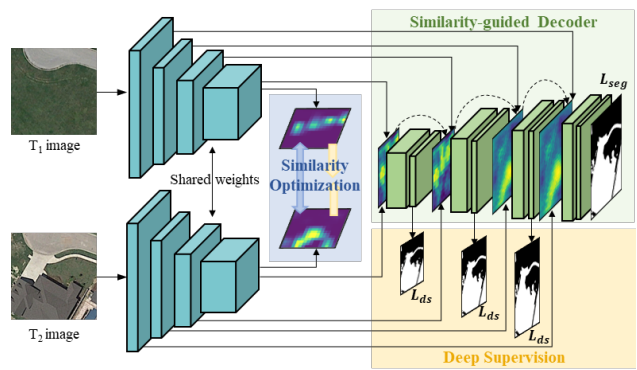

Fig.3. The architecture of our proposed SAAN framework.

*3) Attention Mechanisms for Change Detection*

As the features extracted from the encoder network contain information about the entire image, it is challenging to identify the change regions from redundant background information. To suppress irrelevant information, attention mechanisms have been introduced in CD methods to enable the model to selectively focus on discriminative regions. The effectiveness of attention mechanisms has been demonstrated in both segmentation-based and metric learning-based CD approaches[32], [51]. For instance, Shi et al. [32] incorporated the convolutional block attention module[52] to emphasize informative channels and regions before measuring feature differences. Guo et al.[39] introduced self-attention mechanisms to capture temporal feature dependencies and thereby enhance image feature representations. Furthermore, Cheng et al.[36] designed a multilayer and multi-head self-attention module to capture long-range and hierarchical context features. While the application of attention mechanisms designed for natural images can lead to some improvement in CD accuracy [32], [51], further enhancements can be achieved by considering the specific characteristics of the CD task.

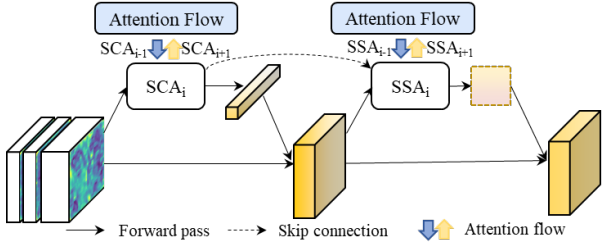

Fig.4. The architecture of the similarity-guided attention flow module.

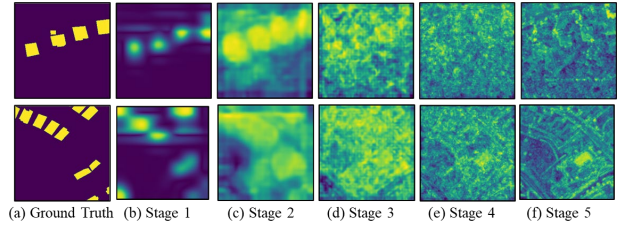

Fig.5. The similarity relation between bi-temporal encoder features deteriorates from lower stage to higher stage.

## III. METHODOLOGY

### A. Method overview

As illustrated in Fig.1, the existing Siamese encoder-decoder architecture cannot effectively exploit informative features from deep encoder layers, resulting in low activation values for change regions in the early decoder stages. Moreover, the model's confidence in the change areas is inconsistent across different decoder stages, indicating that the model is affected by irrelevant non-changing noises contained in the encoder features. To address these challenges, we propose a similarity-aware attention flow network (SAAN) for effective and efficient CD. In the proposed SAAN, we tackle the first problem by introducing similarity optimization in the encoder part to explicitly guide deep convolutional layers in exploiting semantic information from the input images. Moreover, since many segmentation-based models operate as black-box techniques that are optimized in the output space, they lack explicit optimization rules on semantic correlations. To alleviate the second problem, a similarity-guided attention module is designed to enhance change feature representation and enable the model to focus on the changing areas based on the guidance of the similarity relation between bi-temporal features. Furthermore, noting that shallow encoder features primarily capture general information about the entire image and may not adequately capture similarity correlations, we introduce the attention flow mechanism to ensure that the feature similarity correlation can improve change feature representation at all decoder stages.

The architecture of the proposed SAAN framework is shown in Fig.3, which includes a Siamese encoder network, a similarity-guided decoder, and a deep supervision module. The bi-temporal remote sensing images are inputted into the Siamese encoder network for hierarchical feature extraction. A similarity optimization strategy is introduced to ensure that the extracted features are semantically similar in the non-changing areas and semantically dissimilar in the changing areas. In addition to this similarity optimization technique, we further utilize the similarity as guidance within the decoder network by designing a similarity-guided channel-spatial attention mechanism. The similarity-guided channel attention facilitates the fusion of bi-temporal features with deep decoder features by suppressing irrelevant channels, while the spatial attention module further enhances change feature representation in the spatial dimension. To address the feature similarity inconsistency problem across different stages, we adopt an attention flow mechanism to enable the propagation of similarity relations at different decoder stages, thereby preserving the model's perception of feature similarity relationships. Deep supervision is also introduced to improve change feature representation in the decoder stage.

### B. Similarity-optimized Siamese Encoder

VHR remote sensing images provide abundant details of the land surface. Change objects of interest are usually mixed with complex pseudo changes induced by variations in illumination, imaging conditions, and seasons, showing multiscale and multitype characteristics. Many existing segmentation-based methods employ popular frameworks such as VGG-Net[16] and Res-Net[14], to extract bi-temporal image features. However, these methods are optimized based on the loss calculated at the highest decoder stage. Consequently, the encoder part cannot be fully transferred to the CD task due to the vanishing gradient phenomenon, resulting in the deterioration of the change feature representations. Inspired by the metric learning methods[40] that explicitly optimize feature similarity, we aim to incorporate this concept into the segmentation-based methods.

Given the input bi-temporal images $I_{t1}$ and $I_{t2}$, they are processed through the Siamese encoder network to extract bi-temporal image features. The encoder network is optimized by calculating the contrastive loss[40] between bi-temporal encoder features $f_{t1}$ and $f_{t2}$ at the deepest scale. In contrast to many existing metric learning-based methods that directly optimize the Euclidean distance between features, we project feature $f_{t1}$ and $f_{t2}$ into the L2 hypersphere embedding. This helps mitigate the influence of pseudo changes arising from variations in illumination and imaging conditions. The Euclidean distance between the normalized features is equal to the cosine similarity distance between bi-temporal features:

$$d = \sqrt{(\|f_{t1}\|_2 - \|f_{t2}\|_2)^2} = \sqrt{2 - 2\cos(f_{t1}, f_{t2})}, \quad (1)$$

where $d$ denotes the cosine distance between the extracted bi-temporal features. $d = 0$ represents that $f_{t1}$ and $f_{t2}$ are semantically aligned since $\cos(f_{t1}, f_{t2})=1$; $d = \sqrt{2}$ indicates that features are semantically dissimilar in the feature space because $\cos(f_{t1}, f_{t2})=0$. Intuitively, we expect the bi-temporal feature of the non-changing objects to be close in the feature space since the feature representations of the same ground object should be identical. To this end, the cosine distance between the bi-temporal features of the non-changing regions should be optimized towards 0. On the contrary, the bi-temporal features of the changing regions are expected to be semantically different in the feature space. However, directly optimizing the cosine distance towards $\sqrt{2}$ is equivalent to optimizing the features of different categories until they are totally linearly independent, which ignores the potential association between different categories. To tackle this problem, we introduce the margin $m$ to cut off the gradient when the cosine distance of the changed pair exceeds $m$. With the annotated ground truth $y$ and the cosine distance $d$, the contrastive loss can be formulated as:

$$L_{con} = \sum_{i,j} \frac{1}{2}(1-y_{i,j})d_{i,j}^2 + \frac{1}{2}y_{i,j}max(m-d,0)^2, \quad (2)$$

where $m$ is set to 1 as default, indicating that the model optimizes the dissimilarity of different objects until the angle between bi-temporal features is greater than 60 degrees. By using $L_{con}$ as an auxiliary optimization to the network, the encoder network can extract informative semantic features in the deep feature space, and thus avoid ineffective deep feature learning in the encoder stage.

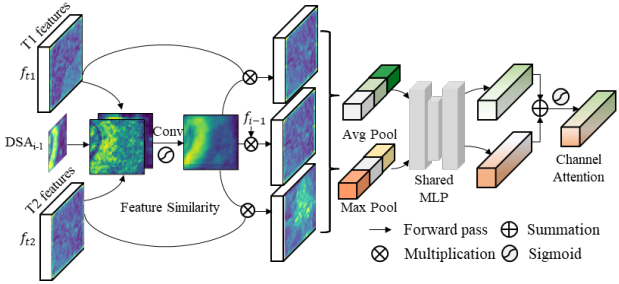

Fig.6. The architecture of Similarity-guided Channel Attention (SCA) module.

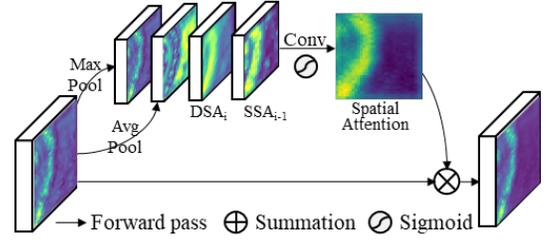

Fig.7. The architecture of Similarity-guided Spatial Attention (SSA) module.

### C. Similarity-guided Decoder

In the decoder part, we design a similarity-guided attention flow module to enhance change feature representation, as shown in Fig.4. Many existing methods directly fuse bi-temporal encoder features and deep decoder features using simple concatenation and convolution. As the encoder features contain general information about the whole image, irrelevant information in the non-changing regions may deteriorate the change feature representation. As a result, the model's confidence in the changing region is inconsistent due to the impact of irrelevant noise. To tackle this problem, we introduce feature similarity relation as guidance to enhance change feature representation in the decoder stage. Specifically, the features are fed into the similarity-guided channel attention (SCA) module to enhance the informative channels and suppress the irrelevant ones. The enhanced features are then fused using two successive convolution modules, each of which includes a convolutional layer of kernel size (3,3), a batch normalization layer, and an activation layer. The fusion results are fed into the similarity-guided spatial attention (SSA) module to further enhance change feature representation in the spatial dimension. Furthermore, we design an attention flow mechanism to enable similarity information flow at different decoder stages. In the subsections below, we elaborate on the details of the proposed components.

*1) Attention Flow Mechanism*

We intend to enhance change feature representation in the decoder stage by designing an attention mechanism guided by feature similarity. However, in Fig. 5, we empirically observe that the similarity relation between shallow convolutional features deteriorates as the spatial resolution increases. This is because shallow encoder features of high-resolution contain general information of the whole image, and the similarity between bi-temporal shallow features cannot well present the semantic changing/non-changing information. Besides, shallow encoder features contain local-detailed shape or texture information that does not help provide localization of the change region in the attention map. To this end, we introduce connection operations to transmit the similarity attention map generated by the previous stage to the current stage. By interconnecting attention modules between adjacent stages, we can ensure that the semantic similarity information can be preserved at different feature resolutions. This is implemented by fusing the similarity attention map transmitted from the attention modules of previous decoder stages. This attention flow design enables the attention module to generate effective attention maps by considering the similarity of both the current and previous decoder stages. In this way, the model performance can be improved by perceiving optimal similarity attention maps at all decoder stages.

*2) Similarity-guided Channel Attention*

We design a similarity-guided channel attention (SCA) mechanism to exploit the inter-channel relationship of features, as illustrated in Fig.6. The similarity-guided channel attention module takes the bitemporal features $f_{t1}$ and $f_{t2}$ transmitted from the Siamese encoder, and the output feature $f_{i-1}$ from the previous decoder module as the input. We leverage the similarity relationship between bi-temporal image features as guidance to exploit the changing information. The cosine similarity between the bi-temporal images can be calculated as follow:

$$Sim_i = sum_c(\|f_{t1}\| \cdot \|f_{t2}\|), \quad (3)$$

where $\|\cdot\|$ denoted the L2 normalization along the channel axis, and $sum_c(\cdot)$ is the summation operation along the channel axis. In the similarity map, values in the non-changing areas are close to 1 since they contain similar feature representations, while values in the changing areas are close to 0 since they are semantically different in the feature space. To alleviate the abovementioned similarity inconsistency problem, an attention flow mechanism is proposed to transmit the attention map from the previous SCA module to the current stage and fuse the transmitted attention map with $Sim_i$ to generate the attention map of the current stage:

$$DSA_i = \sigma(C_{7\times7}^1([Sim_i, DSA_{i-1}])), \quad (4)$$

where $C_{7\times7}^1$ is a convolutional layer of kernel size (7,7) with one output channel; the sigmoid operation $\sigma(\cdot)$ generates a pixel-wise attention map ranging from 0 to 1. $DSA_i$ is multiplied with $f_{t1}$, $f_{t2}$, and $f_{i-1}$ respectively to suppress the non-changing regions and highlight the changing regions. The enhanced features are then concatenated and aggregated using global max pooling and average pooling operations. The features are fed into a parameter-sharing multilayer perception module with one hidden layer. The output vectors are summed and converted into a channel attention map using a sigmoid operation, as illustrated in Fig.6. The similarity-aware channel attention map is computed as:

$$A_c^i(f_{t1}, f_{t2}, f_{i-1}) = \sigma(MLP(AvgPool(DSA_i \cdot [f_{t1}, f_{t2}, f_{i-1}])) + MLP(MaxPool(DSA_i \cdot [f_{t1}, f_{t2}, f_{i-1}]))), (5)$$

where $AvgPool(\cdot)$ and $MaxPool(\cdot)$ represent global average pooling and global max pooling respectively; $[\cdot]$ denotes feature concatenation and $MLP(\cdot)$ denotes the multilayer perception. With similarity-aware channel attention map $A_c^i$, channels with high activation values are highlighted, while channels with low activation values are suppressed. The channel-enhanced features are fed into double convolution layers, followed by a similarity-guided spatial attention (SSA) module that further enhance change feature representations in the spatial dimension.

*3) Similarity-guided Spatial Attention*

The SSA mechanism focuses on 'where' the change areas are by measuring the inter-spatial relationship of features under the guidance of feature similarity, as illustrated in Fig.7. Inspired by the CBAM attention mechanism[52], SSA first aggregates channel-wise information by calculating the mean and max value of each pixel in the feature map. Apart from aggregating information from the feature itself, the proposed attention flow mechanism enables inter-stage similarity information preservation and transmits the spatial attention map $A_s^{i-1}$ generated by stage $i-1$ to the current stage $i$. We also introduce $DSA_i$ generated from the similarity-guided channel attention module as auxiliary information for spatial attention map generation in SSA. Overall, the similarity-guided spatial attention map $A_s^i$ can be generated as follows:

$$A_s^i = \sigma(C_{7\times7}^1([AvgPool(f_i), MaxPool(f_i), A_s^{i-1}, DSA_i]) \quad (6)$$

where $A_s^i$ means the spatial attention map generated at the current stage; $A_s^{i-1}$ represents the spatial attention map transmitted from the previous SSA module through attention flow mechanism. We concatenate and encode the information by applying a convolution kernel of size (7,7) to generate a spatial attention map using a sigmoid operation. The generated spatial attention map $A_s^i$ is multiplied with $f_i$ in the spatial dimension to enhance the changing regions and suppress the non-changing ones. Unlike many existing spatial attention mechanisms that generate spatial attention from the output feature itself, SSA introduces the spatial attention map from the previous stage and the dissimilarity map as auxiliary information to guide the model to focus on the changing areas.

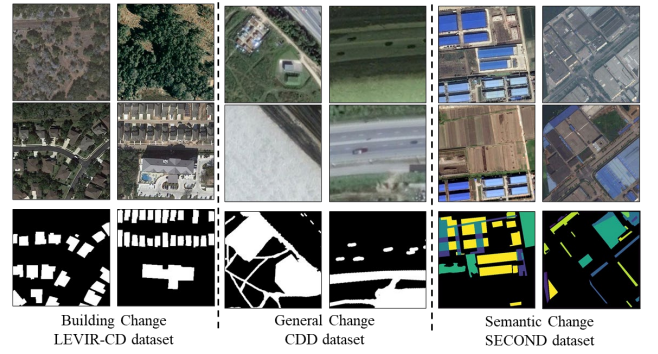

Fig.8. Illustration of the datasets used in this paper.

*4) Relation to the Existing Attention Mechanisms*

Here, we discuss the relations and distinctions between the proposed similarity-guided attention and the existing attention mechanisms. In comparison to the attention mechanisms previously employed for CD[32], the proposed similarity-guided attention module exhibits greater capability in harnessing informative features. This is achieved by leveraging the similarity relation between bi-temporal images to effectively suppress non-changing regions and channels. As a result, the proposed attention module excels in exploiting inter-channel and inter-pixel relations, thereby enabling the model's focus on the changing areas. Although previous studies have incorporated similarity relation into the attention mechanism within domains such as object counting[53] and medical image segmentation [54], the existing similarity-aware mechanisms mainly focus on similarity relation among pixels or channels in the spatial-channel dimension. In contrast, the proposed similarity-guided attention module considers the specific characteristics of the CD task. By capturing the similarity relation between bi-temporal image features, the module guides the model's focus toward the changing regions. Moreover, as SAAN explicitly optimizes feature similarity relation in the feature space, the extracted features are semantically similar in the unchanged regions while exhibiting differences in the changing regions. Another advantage of the proposed method is the attention flow mechanism. Distinct from existing similarity-aware attention mechanisms that solely consider the similarity relation within the current stage, the proposed attention flow interconnects attention modules between adjacent stages, ensuring the model's perception of semantic similarity information across different decoder stages.

TABLE I
CHANGE DETECTION RESULTS ON THE LEVIR-CD DATASET

| Method | Accuracy | | | | Complexity | | |
|---|---|---|---|---|---|---|---|
| | Pre. | Rec. | F1 | IoU | MPara. | GFLOPs | Time(ms) |
| *FC-EF* | 86.61 | 82.85 | 84.69 | 73.44 | 1.34 | 3.55 | 7.43 |
| *FC-SD* | 89.6 | 86.79 | 88.17 | 78.85 | 1.34 | 4.68 | 8.48 |
| *FC-SC* | 89.64 | 87.5 | 88.56 | 79.46 | 1.54 | 5.288 | 8.72 |
| *DSIFN* | 91.92 | 89.65 | 90.77 | 83.11 | 50.44 | 82.25 | 26.97 |
| *DDCNN* | 91.81 | 89.09 | 90.43 | 82.54 | 46.67 | 177.43 | 44.11 |
| *DSAMNet* | 86.26 | 90.13 | 88.15 | 78.82 | 16.95 | 75.35 | 19.93 |
| *SNUNet* | 91.49 | <u>90.26</u> | 90.87 | 83.27 | 12.03 | 54.83 | 28.77 |
| *MSPSNet* | 92.07 | 90.17 | <u>91.11</u> | <u>83.66</u> | 2.21 | 15.04 | 21.17 |
| *BIT* | **92.88** | 87.63 | 90.18 | 82.12 | 3.49 | 21.26 | 45.1 |
| *ChangeFormer* | 92.8 | 88.37 | 90.53 | 82.7 | 41.02 | 405.57 | 65.02 |
| *PA-Former* | 91.7 | 89.31 | 90.49 | 82.63 | 16.13 | 21.71 | 51.79 |
| *SAAN* | <u>92.19</u> | **90.64** | **91.41** | **84.18** | 17.57 | 10.42 | 20.34 |

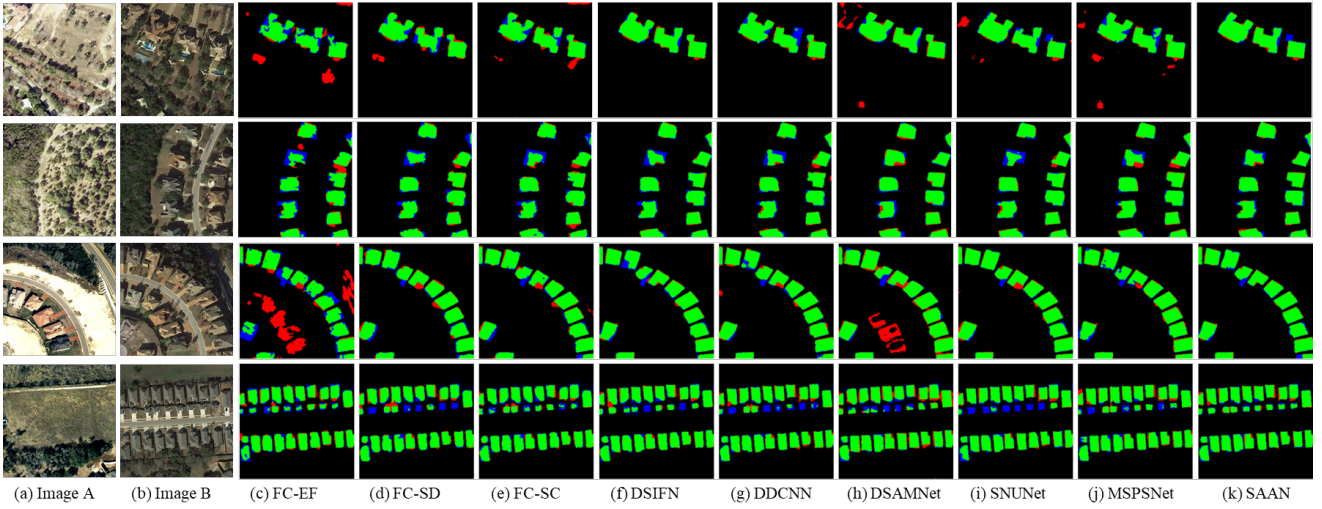

Fig.9. Visualization of building change detection results of the compared methods on the LEVIR-CD dataset. Green represents the pixels that are correctly predicted; red represents misclassified background pixels; blue represents the omitted building pixels.

### D. Optimization

The optimization of SAAN includes three components: the contrastive loss, the deep supervision loss, and the output loss. The contrastive loss $L_{con}$, introduced in Eq. 3, can explicitly guide deep convolutional layers to exploit semantic similarity information from the input images. Then in the decoder part, we introduce deep supervision loss at each decoder stage to guide the decoder to perceive change information in each decoder stage, which also helps alleviate the irrelevant noise problem. Given output feature $f_i$ from the decoder of stage $i$, we introduce a convolutional layer to generate change prediction from $f_i$, and construct auxiliary tasks to optimize the decoder network as follows:

$$L_{aux}^i = Dice(C_{1\times 1}^1(f_i), Down(y)) + CE\left(C_{1\times 1}^1(f_i), Down(y)\right), \quad (7)$$

where $Dice(\cdot)$ denotes the dice loss that alleviates the unbalanced data problem and $CE(\cdot)$ is the cross-entropy loss. We also calculate dice loss and cross-entropy loss at the final model output, which has the same spatial resolution as the input image:

$$L_{seg} = Dice(\hat{y}, y) + CE(\hat{y}, y), \quad (8)$$

where $\hat{y}$ denotes the output prediction of the final model output. The overall loss function is the weighted sum of $L_{seg}$, $L_{con}$ and $L_{aux}$:

$$L = L_{seg} + w \cdot L_{con} + w \cdot \sum_{i=0}^{4} L_{aux}^i. \quad (9)$$

Since $L_{con}$ and $L_{aux}$ are auxiliary tasks that assist model learning, we set $w = 0.3$ to balance between the main task and the auxiliary tasks.

## IV. EXPERIMENTS

### A. Implementation Details

#### 1) Datasets

To validate the effectiveness of the proposed methods, we conducted experiments on a wide range of CD benchmarks, including the LEVIR-CD building CD dataset, the CDD general CD dataset, and the SECOND semantic CD dataset. These

TABLE II
CHANGE DETECTION RESULTS ON THE CDD DATASET

| Method | Accuracy | | | | Complexity | | |
|---|---|---|---|---|---|---|---|
| | Pre. | Rec. | F1 | IoU | MPara. | GFLOPs | Time(ms) |
| FC-EF | 92.32 | 85.7 | 88.89 | 79.99 | 1.34 | 3.55 | 7.28 |
| FC-SD | 94.29 | 92.07 | 93.17 | 87.21 | 1.34 | 4.68 | 8.41 |
| FC-SC | 93.48 | 89.89 | 91.65 | 84.59 | 1.54 | 5.288 | 8.82 |
| DSIFN | 96.82 | 92.8 | 94.77 | 90.05 | 50.71 | 82.25 | 27.73 |
| DDCNN | 94.47 | 94.43 | 94.45 | 89.48 | 46.67 | 177.43 | 43.08 |
| DSAMNet | 92.1 | 93.25 | 92.67 | 86.34 | 16.95 | 75.35 | 19.21 |
| SNUNet | 96.41 | 94.58 | 95.48 | 91.36 | 12.03 | 54.83 | 28.98 |
| MSPSNet | 96.01 | 95 | 95.5 | 91.4 | 2.21 | 15.04 | 20.91 |
| BIT | 96.00 | 94.04 | 95.01 | 90.50 | 3.49 | 21.26 | 45.8 |
| ChangeFormer | 95.01 | 94.27 | 94.64 | 89.82 | 41.02 | 405.57 | 63.51 |
| PA-Former | 94.46 | 92.81 | 93.63 | 88.02 | 16.13 | 21.71 | 52.38 |
| SAAN | **97.81** | **96.27** | **97.03** | **94.23** | 17.57 | 10.42 | 20.81 |

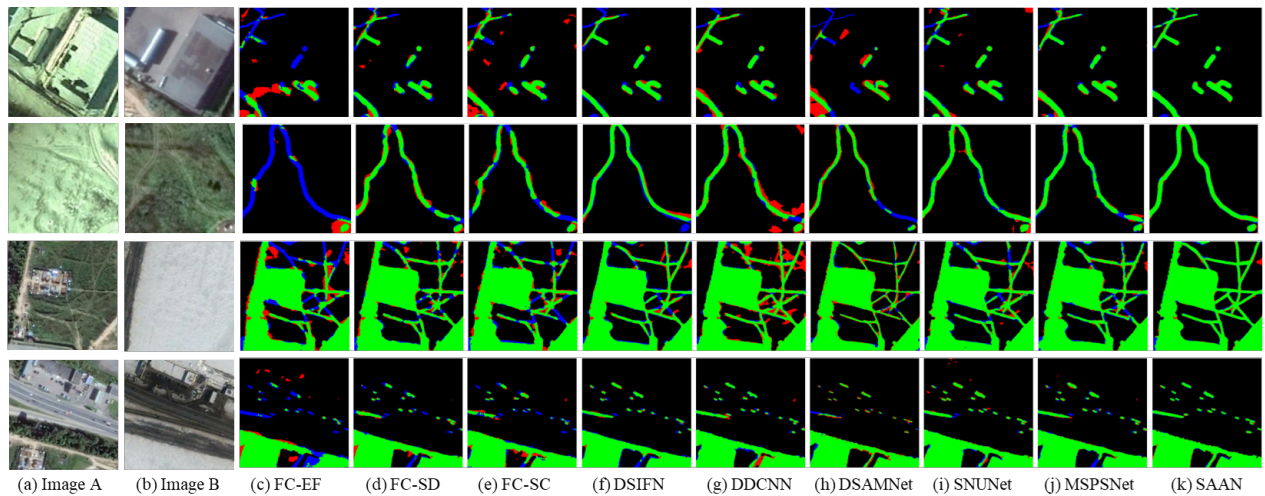

(a) Image A  (b) Image B  (c) FC-EF  (d) FC-SD  (e) FC-SC  (f) DSIFN  (g) DDCNN  (h) DSAMNet  (i) SNUNet  (j) MSPSNet  (k) SAAN

Fig.10. Visualization of genral change detection results of the compared methods on the CDD dataset. Green represents the pixels that are correctly predicted; red represents misclassified background pixels; blue represents the omitted pixels.

datasets, as illustrated in Fig.8, correspond to three fundamental CD tasks respectively and cover different circumstances of sensors, cities, and spatial resolutions. The details of these datasets are described as follows.

1) LEVIR-CD dataset: The LEVIR-CD dataset is a large-scale building CD dataset that covers 20 different cities in the US[48]. The dataset includes images of 0.5m spatial resolution with more than 30,000 change buildings. Due to different imaging times in different regions, the seasonal and illumination changes pose a challenge for the model to exploit real building changes from the irrelevant changes. Following the data processing method in [36], we cropped the dataset into 256x256 image pairs and removed the ones without changing pixels. Consequently, the resulting dataset consists of 3,167, 436, and 935 pairs for training, validation, and testing, respectively. Examples of the LEVIR-CD dataset are shown in Fig.8.

2) CDD dataset: The CDD dataset contains 16,000 season-varying image pairs of spatial resolution ranging from 0.03m to 1m[55]. This dataset includes general change objects of diverse sizes, such as cars, roads, and buildings. Following the provided data partitions, the dataset includes 10000 pairs for the training set, 2998 pairs for the validation set, and 3000 pairs for the testing set. Examples of CDD dataset is shown in Fig.8.

3) SECOND dataset: The SECOND dataset is a semantic change detection benchmark proposed by Yang et al.[56]. The dataset contains 30 types of semantic change information across 6 land-cover categories. The dataset contains 4662 image pairs of size 512x512. The dataset was randomly divided into training, validation, and test sets with a ratio of 7:1:2. Examples of the SECOND dataset are shown in Fig.8.

2) Evaluation Metrics

In the binary change detection tasks, we adopted 4 indication metrics, including precision, recall, F1-score, and IoU to evaluate the effectiveness of the methods. Given the binary change prediction result and its corresponding ground truth, we count the number of true positive (TP), false positive (FP), false negative (FN), and true negative (TN) pixels, and calculate the evaluation metrics as follows:

$$Precision = \frac{TP}{TP+FP} \quad (10)$$

$$Recall = \frac{TP}{TP+FN} \quad (11)$$

$$F1 = \frac{2 \times precision \times recall}{precision+recall} \quad (12)$$

$$IoU = \frac{TP}{TP+FN+FP} \quad (13)$$

As for the semantic CD task, we adopted the IoU, Separated Kappa(SeK) coefficient, and the weighted sum of IoU and SeK to evaluate the semantic change detection performance, as designed by Yang et al.[56]. Please refer to reference [56] for the detailed calculation process of the evaluation metrics adopted for semantic CD.

3) Implementation

The proposed SAAN was implemented by the Pytorch library[57] and was trained on an RTX 3090 GPU with a batch size of twelve. The network was optimized by the Adam optimizer[58] with an initial learning rate of 0.0005 and a weight decay of 1e-5. The learning rate was reduced to one-third of the current learning rate if the validation accuracy failed to increase for 5 epochs. To prevent overfitting, we employed various data augmentation strategies, including random rotation, horizontal flip, and vertical flip. The training process stopped when the current learning rate was below 1e-7. After the model converged, we evaluated the model's performance on the test dataset by calculating the evaluation metrics based on the CD predictions.

TABLE III
SEMANTIC CHANGE DETECTION RESULTS ON THE SECOND DATASET

| Methods | Accuracy | | | Complexity | | |
|---|---|---|---|---|---|---|
| | IoU | SeK | Overall Score | MPara. | GFLOPs | Time(ms) |
| HRSCD-S1 | 32.78 | 7.19 | 14.86 | 6.03 | 13.61 | 21.53 |
| HRSCD-S2 | 35.37 | 6.74 | 15.32 | 6.03 | 7.03 | 21.39 |
| HRSCD-S3 | 38.77 | 7.73 | 17.04 | 15.63 | 14.52 | 23.04 |
| HRSCD-S4 | 50.85 | 15.2 | 25.89 | 15.83 | 15.11 | 24.16 |
| ChangeMask | <u>53.35</u> | <u>17.24</u> | <u>28.07</u> | 10.62 | 18.96 | 48.6 |
| SAAN | **53.49** | **18.03** | **28.66** | 17.53 | 15.79 | 30.18 |

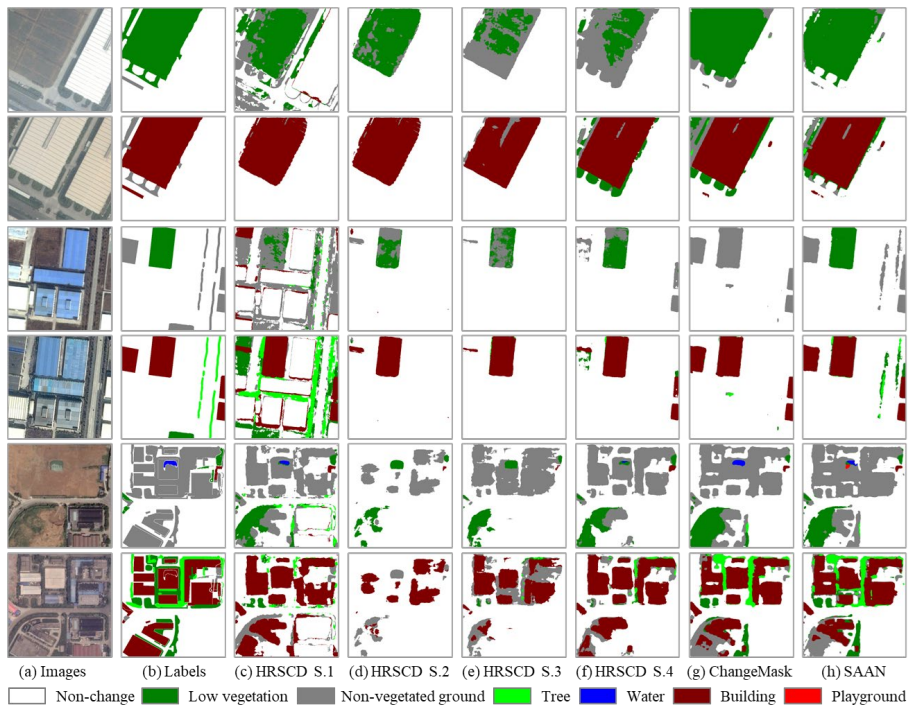

Fig.11. Visualization of semantic change detection results of the compared methods on the SECOND dataset.

### B. Results on LEVIR Dataset

Table I reports the quantitative evaluation results of the proposed SAAN and several comparison methods on the test set of the LEVIR-CD dataset. The highest scores are highlighted in bold, while the second-highest results are underlined. In the binary change detection task, our method is compared with recent works in binary CD, including FC-EF[31], FC-SD[31], FC-SC[31], DSIFN[32], DDCNN[59], DSAMNet[51], SNUNet[60], MSPSNet[39], BIT[61], ChangeFormer[62], and PA-Former[63]. We also report the complexity of the comparison methods, including the number of parameters(MPara.), the number of floating point operations(GFLOPs), and the computational time per iteration in the test phase(Time(ms)). The quantitative evaluation results in Table I show that our SAAN method, incorporating similarity optimization and similarity-guided attention flow mechanism, surpasses the current state-of-the-art CD methods. SAAN outperformed the competitors by 91.41% on the F1 score and 84.18% on IoU, respectively. Compared to the second-best method, MSPSNet, SAAN increased the F1-score and IoU by 0.3% and 0.52% on the LEVIR-CD dataset, respectively. SAAN has a relatively fast inference speed and low GFLOPs compared to the comparison methods. SAAN also surpasses the state-of-the-art transformer-based CD methods in terms of accuracy and computational complexity.

In Fig.9, we visualize the CD prediction results obtained from SAAN and the comparison methods. It can be observed from Fig.9 rows 1-3 that SAAN precisely discriminates the changing regions from the background and exhibits minimal misclassifications in the non-changing areas, which indicates that SAAN successfully suppresses pseudo changes due to the effectiveness of the proposed similarity-guided attention module. In comparison, some of the pseudo-seasonal changes and non-changing buildings are misclassified by FC-EF, DSAMNet, and MSPSNet. Moreover, we can see from Fig.9 row 4 that the small buildings are omitted by all the comparison methods. SAAN, however, precisely detects the small buildings without omission. This performance enhancement is attributed to the attention flow mechanism, which preserves the model's perception of feature similarity across different decoder stages and improves performance on spatial details.

TABLE IV
ABLATION EXPERIMENTS ON THE OPTIMIZATION STRATEGIES

| Model name | LEVIR | | CDD | | SECOND | |
|---|---|---|---|---|---|---|
| | F1(%) | IoU(%) | F1(%) | IoU(%) | SeK(%) | IoU(%) |
| Opt-A: SiamRes18 | $90.27_{\pm 0.04}$ | $82.28_{\pm 0.06}$ | $96.29_{\pm 0.01}$ | $92.85_{\pm 0.2}$ | $17.34_{\pm 0.02}$ | $53.05_{\pm 0.02}$ |
| Opt-B: SiamRes18+Sim | $90.64_{\pm 0.04}$ | $83.05_{\pm 0.07}$ | $96.65_{\pm 0.02}$ | $93.51_{\pm 0.04}$ | $17.49_{\pm 0.01}$ | $53.22_{\pm 0.02}$ |
| Opt-C: SiamRes18+DS | $90.86_{\pm 0.05}$ | $83.22_{\pm 0.08}$ | $96.79_{\pm 0.04}$ | $93.79_{\pm 0.08}$ | $17.48_{\pm 0.06}$ | $53.02_{\pm 0.02}$ |
| Opt-D: SiamRes18+Sim+DS | **$90.91_{\pm 0.02}$** | **$83.47_{\pm 0.04}$** | **$96.85_{\pm 0.01}$** | **$93.89_{\pm 0.3}$** | **$17.6_{\pm 0.07}$** | **$53.24_{\pm 0.05}$** |

TABLE V
ABLATION EXPERIMENTS ON THE ATTENTION MECHANISMS

| Model name | Channel Attention | Spatial Attention | Attention Flow | LEVIR F1 | LEVIR IoU | CDD F1 | CDD IoU | SECOND SeK | SECOND IoU |
|---|---|---|---|---|---|---|---|---|---|
| Opt-D | | | | $90.91_{\pm0.06}$ | $83.47_{\pm0.09}$ | $96.85_{\pm0.01}$ | $93.89_{\pm0.02}$ | $17.6_{\pm0.04}$ | $53.24_{\pm0.05}$ |
| Opt-D+SCA | √ | | | $91.1_{\pm0.04}$ | $83.66_{\pm0.06}$ | $96.86_{\pm0.01}$ | $93.92_{\pm0.01}$ | $17.65_{\pm0.03}$ | $53.38_{\pm0.04}$ |
| Opt-D+SCA+ flow | √ | | √ | $91.33_{\pm0.03}$ | $84.05_{\pm0.04}$ | $96.94_{\pm0.04}$ | $94.07_{\pm0.06}$ | $17.84_{\pm0.06}$ | $53.53_{\pm0.05}$ |
| Opt-D+SSA | | √ | | $91.01_{\pm0.05}$ | $83.52_{\pm0.08}$ | $96.93_{\pm0.03}$ | $94.05_{\pm0.05}$ | $17.82_{\pm0.06}$ | $53.31_{\pm0.09}$ |
| Opt-D+SSA+ flow | | √ | √ | $91.29_{\pm0.08}$ | $83.98_{\pm0.12}$ | $96.97_{\pm0.06}$ | $94.11_{\pm0.09}$ | $17.98_{\pm0.03}$ | $53.57_{\pm0.05}$ |
| Opt-D+SCA+SSA | √ | √ | | $91.24_{\pm0.03}$ | $83.89_{\pm0.05}$ | $96.95_{\pm0.02}$ | $94.08_{\pm0.03}$ | $17.96_{\pm0.09}$ | $53.4_{\pm0.08}$ |
| Opt-D+SCA+SSA+flow | √ | √ | √ | $91.39_{\pm0.07}$ | $84.15_{\pm0.13}$ | $97.03_{\pm0.04}$ | $94.23_{\pm0.06}$ | $18.03_{\pm0.02}$ | $53.49_{\pm0.07}$ |

## C. Results on CDD Dataset

We conducted experiments on the CDD dataset to further verify the effectiveness of SAAN on the general CD task, which poses a greater challenge for the model to discriminate real change regions of varied sizes from massive pseudo-seasonal changes. The quantitative experimental results are presented in Table II, demonstrating the remarkable superiority of SAAN over the current state-of-the-art CD methods. Our proposed SAAN achieved 97.03% on the F1 score and 94.23% on IoU, surpassing the currently best segmentation-based MSPSNet and metric learning-based DSAMNet by 1.53% and 3.78% on the F1 score, respectively.

We visualize the CD predictions generated by SAAN and the comparison methods on the CDD dataset in Fig. 10. We can observe from Fig.10 that our proposed SAAN successfully identifies real changes from complex backgrounds. Rows 1-3 exhibit examples where FC-EF, DDCNN, and DSAMNet fail to accurately discriminate changed roads and buildings, resulting in significant misclassifications and omissions in the CD prediction outputs. Additionally, the changing small vehicles in Fig. 10 row 4 are omitted by most of the comparison methods. In comparison, the predictions generated by SAAN completely cover the changing regions with the least misclassifications and omissions, thereby demonstrating the efficacy of our proposed method.

## D. Results on SECOND Dataset

To validate the generalization of the proposed SAAN framework, we conducted experiments on the semantic change detection SECOND dataset, which requires not only the binary CD results but also semantic change information in the form of 'from-to' transitions from bi-temporal images. The quantitative evaluation results are presented in Table III. Our method is compared with five related semantic change detection methods including ChangeMask[27] and HRSCD[64] strategies 1-4. To ensure a fair comparison with SAAN, HRSCD[64] strategies 1-

TABLE VI
COMPUTATIONAL COST OF THE PROPOSED COMPONENTS

| Model name | Binary Change Detection | | | Semantic Change Detection | | |
|---|---|---|---|---|---|---|
| | ΔMPara. | ΔGFLOPs | ΔTime(ms) | ΔMPara. | ΔGFLOPs | ΔTime(ms) |
| Opt-D +SCA | 0.4057 | 0.0015 | 3.51 | 0.1292 | 0.0013 | 6.18 |
| Opt-D + SCA+flow | 0.4059 | 0.0032 | 4.20 | 0.1294 | 0.0026 | 7.13 |
| Opt-D + SSA | 0.0008 | 0.0011 | 2.44 | 0.0007 | 0.0011 | 1.42 |
| Opt-D + SSA+flow | 0.0010 | 0.0017 | 2.64 | 0.001 | 0.0017 | 1.56 |
| Opt-D +SCA+SSA | 0.4063 | 0.0023 | 4.22 | 0.1298 | 0.0021 | 7.08 |
| Opt-D +SCA+SSA+ flow | **0.4068** | **0.0036** | **4.61** | **0.1301** | **0.0037** | **7.26** |

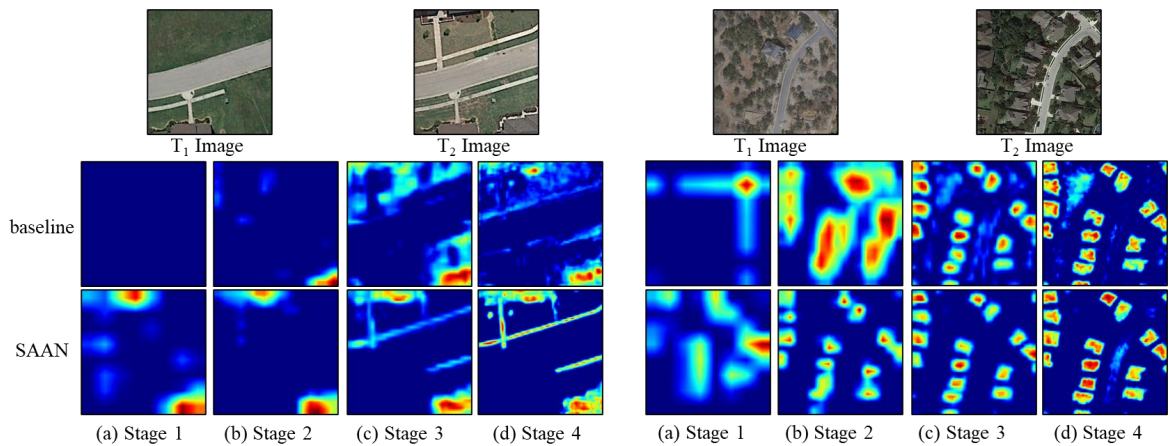

Fig.12. Visualization of the gradient-weighted class activation maps of different decoder stages.

4 were implemented using the ResNet18 backbone. As shown in Table III, SAAN achieves the highest scores in terms of SeK, IoU, and overall score on the SECOND dataset. Compared with the recent ChangeMask[27] method, the Sek and IoU of SAAN are increased by 0.14% and 0.79%, respectively. These results indicate that optimization of feature similarity relationships and utilizing similarity guidance to enhance feature representations can enhance both change region localization and semantic change exploitation. Additionally, SAAN exhibits lower computational costs in terms of GFLOPs.

We visualize some example predictions on the SECOND test dataset in Fig.11, from which we can see that SAAN can accurately locate the changing region and correctly classify the semantic change categories. In Fig.11 rows 1-4, we can see that SAAN can better capture the structure of small changes, which demonstrates the effectiveness of the proposed similarity-guided attention flow mechanism. On the contrary, some detailed changes are either omitted (e.g., HRSCD S.3) or over-segmented (e.g., HRSCD S.1) by the comparison methods. Moreover, in Fig 11 rows 5-6, SAAN can also precisely localize and classify the changing areas in complex urban scenes, affirming the efficacy of the proposed method. Overall, the promising experimental results obtained from building CD, general CD, and semantic CD tasks demonstrate the effectiveness and generalization capabilities of SAAN across various CD applications.

*E. Ablation Study and Analysis*

We conducted ablation experiments on three CD tasks to validate the effectiveness of the proposed components. Firstly, we examined the contributions of the optimization strategies, including the proposed similarity optimization (denoted as Sim) and deep supervision (denoted as DS). The baseline model (denoted as SiamRes18) consists of a Siamese ResNet18 encoder and a U-Net decoder that concatenates and fuses bi-temporal image features using 3x3 convolutions. We performed the experiments five times and reported the mean and variation of the experimental results in Table IV. From Table IV, it can be observed that both similarity optimization and deep supervision improve accuracy across all three CD tasks. The similarity optimization and deep supervision improve the IoU by 0.53% and 0.68% on average in three CD tasks. As a result, combining similarity optimization and deep supervision in CD yields an average IoU improvement of 0.86%.

With similarity optimization and deep supervision, we further conducted ablation experiments on the proposed decoder components, including similarity-guided channel attention, similarity-guided spatial attention, and attention flow mechanism. The ablation experiments results are presented in Table V. Quantitative analysis of the results indicates that both the proposed similarity-guided channel attention and similarity-guided spatial attention modules improve the F1 score and IoU across all three CD tasks. Combining SCA and SSA leads to a linear accumulation of the improvements achieved by each module. This is because SCA utilizes feature similarity relations as guidance to fuse encoder features, while SSA further enhances the fused features by improving the representation of change features in the spatial dimension. Combining SCA and SSA can complementarily fuse the advantages of both modules. Moreover, the attention flow mechanism enables the flow of similarity relations at different decoder stages, thus preserving the model's perception of feature similarity relations and significantly improving the performance of SSA and SCA. As a result, combining SCA, SSA, and attention flow improves the CD accuracy by 0.68%, 0.34% on IoU, and 0.43% on SeK on building CD, general CD, and semantic CD tasks, respectively.

TABLE VII
PERFORMANCE OF DIFFERENT BACKBONE ARCHITECTURE ON THE LEVIR-CD DATASET

| Backbone | Baseline | | SAAN | |
|---|---|---|---|---|
| | F1-Score | IoU | F1-Score | IoU |
| EfficientNet-b0 | 90.46 | 82.58 | 91.38 $_{(+0.92)}$ | 84.13 $_{(+1.55)}$ |
| ResNet18 | 90.27 | 82.28 | 91.39 $_{(+1.12)}$ | 84.15 $_{(+1.87)}$ |
| ResNet34 | 90.7 | 82.99 | 91.16 $_{(+0.46)}$ | 83.75 $_{(+0.76)}$ |
| ResNet50 | 91.39 | 84.15 | 91.81 $_{(+0.42)}$ | 84.87 $_{(+0.72)}$ |
| SE-ResNet50 | 91.23 | 83.88 | 91.58 $_{(+0.35)}$ | 84.47 $_{(+0.59)}$ |

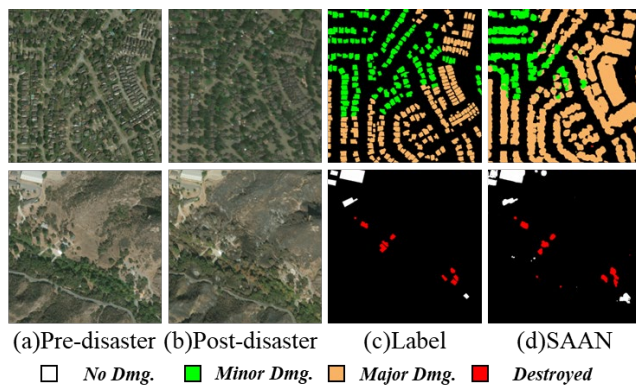

(a)Pre-disaster (b)Post-disaster (c)Label (d)SAAN

☐ *No Dmg.*  ■ *Minor Dmg.*  ■ *Major Dmg.*  ■ *Destroyed*

Fig.13. Qualitative results on the xBD dataset.

TABLE VIII
BENCHMARK COMPARISON ON THE XBD DATASET

| Methods | F1-score | | | | |
|---|---|---|---|---|---|
| | No Dmg. | Minor Dmg. | Major Dmg. | Destroyed | Avg. |
| xView2 Baseline | 66.31 | 14.35 | 0.94 | 46.57 | 32.04 |
| Siamese-UNet | 86.74 | 50.02 | 64.43 | 71.68 | 68.21 |
| ChangeOS | 88.61 | 52.09 | 70.36 | 79.65 | 72.67 |
| SAAN | **91.27** | **53.59** | **71.98** | **80.95** | **74.45** |

We compare the computational cost of the proposed components in Table VI, from which we can see that the proposed SCA, SSA, and attention flow slightly increase the computation costs. Specifically, SCA and SSA bring approximately 0.4M and 0.001M increments in the parameter numbers in the binary CD task; SCA and SSA increase the parameter numbers by 0.13M and 0.0007M in the semantic CD task. It indicates that the proposed components remarkably enhance CD accuracy with minimal additional computational cost. The proposed SAAN is an effective and efficient method for various CD tasks.

In Fig.12, we visualize and compare the class activation maps generated at different decoder stages using Grad-CAM[65]. We can see that the baseline model fails to focus on the changing regions in the early decoder stages, indicating ineffective deep feature learning. With the guidance of feature similarity relation, our proposed SAAN can provide coarse localization of changing regions in the early decoder stage and gradually refine the prediction using shallow encoder features. As a result, our proposed method can generate a higher activation value in the changing regions, leading to improved CD accuracy. This observation highlights the effectiveness of the proposed similarity optimization strategy and the similarity-guided attention flow mechanism.

SAAN can also be extended to other backbone architectures for enhanced CD performance. We apply the proposed similarity optimization and similarity-guided decoder to five different backbone architectures, including EfficientNet[66], ResNet18[14], ResNet34[14], ResNet50[14], and SE-ResNet50[15]. Table VII presents the performance of SAAN across these backbone architectures on the LEVIR-CD dataset. The results demonstrate the general capacity of SAAN. Additionally, the results indicate that lightweight backbone architectures such as EfficientNet-b0 and ResNet18 exhibit a higher improvement in accuracy, which suggests that superior feature representation can be achieved with the explicit guidance of feature similarity.

### F. Application to Building Damage Assessment

To assess the generalization ability of SAAN in CD applications, we apply it to the xBD dataset[2], a large-scale dataset designed for evaluating building damage from satellite imagery. Given satellite images captured during pre-disaster and post-disaster periods, this task aims to detect and classify buildings into different damage statuses, including no damage, minor damage, major damage, and destruction. In SAAN, we adopt a Siamese ResNet-18 encoder with the proposed similarity-guided decoder for building damage assessment. The comparison methods include xView2 Baseline, xView2 1st place solution method Siamese-UNet, and ChangeOS[67]. The models were trained on the Tier3 and training set and subsequently evaluated on the xBD holdout split. Table VIII presents the benchmark comparison results on the xBD holdout split. SAAN outperforms the existing building damage assessment methods in capturing the bi-temporal feature similarity relationship that reflects the degree of building damage. Furthermore, the visualization results in Fig.13 showcase the successful detection of building damage under different disasters using the proposed similarity-guided attention flow module.

## V. CONCLUSION

Inefficient feature learning and inconsistent confidence are two main problems that hinder the prevailing Siamese encoder-decoder architecture from obtaining precise CD results. To address these issues, we introduce a similarity optimization strategy that optimizes deep features to be semantically approximate in the unchanged regions and dissimilar in the changing regions. The optimized feature similarity relation is then utilized as attention to guide the decoder to consistently focus on change regions, thereby alleviating the problem of inconsistent confidence. To this end, we design a lightweight and effective similarity-guided attention module to enhance change feature representation under the guidance of similarity relations. Additionally, we employ an attention flow mechanism to connect adjacent attention blocks at different decoder stages, enabling the seamless propagation of similarity information throughout the decoder. Experimental evaluations demonstrate the superiority of the proposed SAAN framework over recent state-of-the-art methods across various change detection tasks. Importantly, these improvements are achieved with minimal additional computational cost, making our approach highly practical for remote sensing applications.

> REPLACE THIS LINE WITH YOUR PAPER IDENTIFICATION NUMBER (DOUBLE-CLICK HERE TO EDIT) <    14## REFERENCES

[1] M. Decuyper *et al.*, 'Continuous monitoring of forest change dynamics with satellite time series', *Remote Sensing of Environment*, vol. 269, p. 112829, Feb. 2022, doi: 10.1016/j.rse.2021.112829.

[2] R. Gupta *et al.*, 'xBD: A Dataset for Assessing Building Damage from Satellite Imagery', *arXiv:1911.09296 [cs]*, Nov. 2019, Accessed: Sep. 15, 2021. [Online]. Available: http://arxiv.org/abs/1911.09296

[3] A. Alonso-González, C. López-Martínez, K. P. Papathanassiou, and I. Hajnsek, 'Polarimetric SAR Time Series Change Analysis Over Agricultural Areas', *IEEE Transactions on Geoscience and Remote Sensing*, vol. 58, no. 10, pp. 7317–7330, 2020, doi: 10.1109/TGRS.2020.2981929.

[4] A. SINGH, 'Review Article Digital change detection techniques using remotely-sensed data', *International Journal of Remote Sensing*, vol. 10, no. 6, pp. 989–1003, Jun. 1989, doi: 10.1080/01431168908903939.

[5] G. M. Woodwell, J. E. Hobbie, R. A. Houghton, J. M. Melillo, and B. J. Peterson, 'Deforestation measured by LANDSAT: steps toward a method', *Deforestation*, 1983, Accessed: Apr. 15, 2022. [Online]. Available: http://www.researchgate.net/publication/37903366_Deforestation_measured_by_LANDSAT_steps_toward_a_method/amp

[6] S. Liu, L. Bruzzone, F. Bovolo, M. Zanetti, and P. Du, 'Sequential Spectral Change Vector Analysis for Iteratively Discovering and Detecting Multiple Changes in Hyperspectral Images', *IEEE Transactions on Geoscience & Remote Sensing*, vol. 53, no. 8, pp. 4363–4378, 2015, doi: 10.1109/TGRS.2015.2396686.

[7] Allan *et al.*, 'Multivariate Alteration Detection (MAD) and MAF Postprocessing in Multispectral, Bitemporal Image Data: New Approaches to Change Detection Studies', *Remote Sensing of Environment*, 1998, doi: 10.1016/S0034-4257(97)00162-4.

[8] A. A. Nielsen, 'The Regularized Iteratively Reweighted MAD Method for Change Detection in Multi- and Hyperspectral Data', *IEEE Transactions on Image Processing*, vol. 16, p. p.463-478, 2007, doi: 10.1109/TIP.2006.888195.

[9] L. Bruzzone, R. Cossu, and G. Vernazza, 'Detection of land-cover transitions by combining multidate classifiers', *Pattern Recognition Letters*, vol. 25, no. 13, pp. 1491–1500, Oct. 2004, doi: 10.1016/j.patrec.2004.06.002.

[10] Z. Li, C. Yan, Y. Sun, and Q. Xin, 'A Densely Attentive Refinement Network for Change Detection based on Very-High-Resolution Bi-Temporal Remote Sensing Images', *IEEE Transactions on Geoscience and Remote Sensing*, pp. 1–1, 2022, doi: 10.1109/TGRS.2022.3159544.

[11] H. Chen, C. Wu, B. Du, L. Zhang, and L. Wang, 'Change Detection in Multisource VHR Images via Deep Siamese Convolutional Multiple-Layers Recurrent Neural Network', *IEEE Transactions on Geoscience and Remote Sensing*, vol. 58, no. 4, pp. 2848–2864, Apr. 2020, doi: 10.1109/TGRS.2019.2956756.

[12] W. Shi, M. Zhang, R. Zhang, S. Chen, and Z. Zhan, 'Change Detection Based on Artificial Intelligence: State-of-the-Art and Challenges', *Remote Sensing*, vol. 12, no. 10, Art. no. 10, Jan. 2020, doi: 10.3390/rs12101688.

[13] G. Koch, R. Zemel, and R. Salakhutdinov, 'Siamese Neural Networks for One-shot Image Recognition', Accessed: May 17, 2022. [Online]. Available: http://www.cs.utoronto.ca/~rsalakhu/papers/oneshot1.pdf

[14] K. He, X. Zhang, S. Ren, and J. Sun, 'Deep Residual Learning for Image Recognition', *arXiv:1512.03385 [cs]*, Dec. 2015, Accessed: Nov. 05, 2020. [Online]. Available: http://arxiv.org/abs/1512.03385

[15] J. Hu, L. Shen, S. Albanie, G. Sun, and E. Wu, 'Squeeze-and-Excitation Networks', *arXiv:1709.01507 [cs]*, May 2019, Accessed: Mar. 07, 2022. [Online]. Available: http://arxiv.org/abs/1709.01507

[16] K. Simonyan and A. Zisserman, 'Very Deep Convolutional Networks for Large-Scale Image Recognition', *arXiv:1409.1556 [cs]*, Apr. 2015, Accessed: Jun. 03, 2021. [Online]. Available: http://arxiv.org/abs/1409.1556

[17] J. Chen *et al.*, 'DASNet: Dual Attentive Fully Convolutional Siamese Networks for Change Detection in High-Resolution Satellite Images', *IEEE Journal of Selected Topics in Applied Earth Observations and Remote Sensing*, vol. 14, pp. 1194–1206, 2021, doi: 10.1109/JSTARS.2020.3037893.

[18] Z. Zheng, A. Ma, L. Zhang, and Y. Zhong, 'Change is Everywhere: Single-Temporal Supervised Object Change Detection in Remote Sensing Imagery', *arXiv:2108.07002 [cs]*, Aug. 2021, Accessed: Apr. 12, 2022. [Online]. Available: http://arxiv.org/abs/2108.07002

[19] K. Winkler, R. Fuchs, M. Rounsevell, and M. Herold, 'Global land use changes are four times greater than previously estimated', *Nat Commun*, vol. 12, no. 1, Art. no. 1, May 2021, doi: 10.1038/s41467-021-22702-2.

[20] H. Zhang, M. Lin, G. Yang, and L. Zhang, 'ESCNet: An End-to-End Superpixel-Enhanced Change Detection Network for Very-High-Resolution Remote Sensing Images', *IEEE Transactions on Neural Networks and Learning Systems*, pp. 1–15, 2021, doi: 10.1109/TNNLS.2021.3089332.

[21] B. Du, L. Ru, C. Wu, and L. Zhang, 'Unsupervised Deep Slow Feature Analysis for Change Detection in Multi-Temporal Remote Sensing Images', *IEEE Trans. Geosci. Remote Sensing*, vol. 57, no. 12, pp. 9976–9992, Dec. 2019, doi: 10.1109/TGRS.2019.2930682.

[22] R. F. Nelson, 'Detecting forest canopy change using LANDSAT', 1982, doi: http://dx.doi.org/.

[23] J. R. Jensen, 'Urban/suburban land use analysis', *manual of remote sensing second edition*, 1983, Accessed: Apr. 15, 2022. [Online]. Available: